\documentclass[conference]{IEEEtran}
\IEEEoverridecommandlockouts
\usepackage{cite}
\usepackage{amsmath,amssymb,amsfonts}
\usepackage{algorithmic}
\usepackage{graphicx}
\usepackage{textcomp}
\usepackage{xcolor}
\usepackage[ruled,vlined]{algorithm2e}
\def\BibTeX{{\rm B\kern-.05em{\sc i\kern-.025em b}\kern-.08em
    T\kern-.1667em\lower.7ex\hbox{E}\kern-.125emX}}
\begin{document}

\title{Harnessing the Power of Explanations for Incremental Training: A LIME-Based Approach\\

}



    \author{\IEEEauthorblockN{Arnab Neelim Mazumder\IEEEauthorrefmark{1}\IEEEauthorrefmark{2},
    Niall Lyons\IEEEauthorrefmark{1}, Ashutosh Pandey\IEEEauthorrefmark{1}, 
    Avik Santra\IEEEauthorrefmark{1}, and Tinoosh Mohsenin\IEEEauthorrefmark{2}}
    \IEEEauthorblockA{\IEEEauthorrefmark{1}Infineon Technologies, Irvine, CA, USA,\\
    \IEEEauthorrefmark{2}University of Maryland Baltimore County, Baltimore, MD, USA\\
    arnabm1@umbc.edu,
    niall.lyons@infineon.com,
    ashutosh.pandey@infineon.com,\\
    avik.santra@infineon.com, 
    and tinoosh@umbc.edu}}

\maketitle

\begin{abstract}
    Explainability of neural network prediction is essential to understand feature importance and gain interpretable insight into neural network performance. However, explanations of neural network outcomes are mostly limited to visualization, and there is scarce work that looks to use these explanations as feedback to improve model performance. In this work, model explanations are fed back to the feed-forward training to help the model generalize better. To this extent, a custom weighted loss where the weights are generated by considering the Euclidean distances between true LIME (Local Interpretable Model-Agnostic Explanations) explanations and model-predicted LIME explanations is proposed. Also, in practical training scenarios, developing a solution that can help the model learn sequentially without losing information on previous data distribution is imperative due to the unavailability of all the training data at once. Thus, the framework incorporates the custom weighted loss with Elastic Weight Consolidation (EWC) to maintain performance in sequential testing sets. The proposed custom training procedure results in a consistent enhancement of accuracy ranging from 0.5\% to 1.5\% throughout all phases of the incremental learning setup compared to traditional loss-based training methods for the keyword spotting task using the Google Speech Commands dataset. 
\end{abstract}


\section{Introduction}
    Human-Machine interface via voice has become omnipresent in nowadays society. A distinctive feature of voice assistants is that, in order to be used, they first have to be activated by means of a spoken Keyword Spotting (KWS), thereby avoiding computational expenses when it is not required. Thus, KWS can be defined as the task of identification of keywords in audio streams comprising speech and has become a fast-growing technology due to the paradigm shift introduced by deep learning \cite{wwd_review}. The earliest approach is based on continuous KWS. One of the advantages of this approach is the flexibility to deal with changing/non-predefined keywords. Whereas the main disadvantage of such KWS systems might reside in the computational complexity dimension and non-availability of all real-world sequential data at once. 
	
	Therefore it is usually desired to have Incremental Learning (IL) algorithms, commonly referred to as continual learning \cite{cl_principle, cl_har}. In a continual learning setup, a continuously learning agent at a time step "$t$" is trained to recognize the tasks $1, . ., t-1, t$ while the data $\mathcal{D}_1, \mathcal{D}_2, \cdots, \mathcal{D}_{t-1}$, for the tasks $1, \ldots, t-1$ may or may not be available. Such a learning paradigm has two fundamental trade-offs to overcome. The first of these is Knowledge Transfer (KT), which measures how incremental learning up to task $t$ influences the agent's knowledge about it \cite{cl_principle}. In terms of performance, a positive KT suggests that the agent should deliver better accuracy on the task $t$ if allowed to learn it incrementally through tasks $1, \ldots, t-1$ while achieving a low validation error on all of these datasets, assuming that these datasets are seen in the order $\mathcal{D}_1, \mathcal{D}_2, \cdots, \mathcal{D}_t$. On the other hand, Semantic Transfer (ST) measures the influence that learning a task $t$ has on the performance of a previous task \cite{cl_taskSemantic}. A positive ST means that learning a new task $t$ would increase the performance of the model on the previously learned tasks $1, \ldots, t-1$. This compromise between learning a new task and preserving knowledge on previously learned tasks. Multiple methods are proposed in the literature to find the trade-off between KT and ST (architecture-based, memory-based, regularization-based). Here, architectural approaches (e.g., progressive nets \cite{progressive_NN}) evolve the network size after every task while assimilating the new knowledge with the past knowledge into the new network and memory approaches (e.g., gradient episodic memory \cite{gradient_memNN}) store memory of each of the previous tasks \cite{cl_memoryExample} and while learning the new task. In contrast, the regularization method (in this case, elastic weight consolidation) typically assumes a fixed network size and learns a new task while trying to avoid changes to parameters sensitive to previous tasks. 
	
	In this work, a regularization-based Elastic Weight Consolidation (EWC) approach is utilized for continual learning where parameter $\boldsymbol{\theta}_{1: i-1}^*$ configuration is achieved at the end of the dataset $i$, which is expected to solve all the datasets $\mathcal{D}_{1: i}$ \cite{kirkpatrick2017overcoming}. Thus the posterior maximization over the new task is equivalent to likelihood maximization for the new dataset, and the posterior maximization on the previous dataset, 
	\begin{equation}
			\begin{aligned}
			\max _{\boldsymbol{\theta}} \log p\left(\boldsymbol{\theta} \mid \mathcal{D}_{1: i}\right) &=\max _{\boldsymbol{\theta}}\left[\log p\left(\mathcal{D}_{1: i} \mid \boldsymbol{\theta}\right)+\log p(\boldsymbol{\theta})\right] \\
			&=\max _{\boldsymbol{\theta}}\left[\log p\left(\mathcal{D}_i \mid \boldsymbol{\theta}\right)+\log p\left(\boldsymbol{\theta} \mid \mathcal{D}_{1: i-1}\right)\right]
		\end{aligned}
	\label{eq:1}
	\end{equation}

	Such an objective can be minimized by adding a regularization loss, which prevents $\boldsymbol{\theta}_{1: i}^*$ from veering too far away from $\boldsymbol{\theta}_{1: i-1}^*$. Since this regularization loss should preserve closeness to the previous solution, the KL-divergence between $p\left(\boldsymbol{\theta} \mid \mathcal{D}_{1: i}\right)$ and $p\left(\boldsymbol{\theta} \mid \mathcal{D}_{1: i-1}\right)$ as the regularization loss is used. In practice, EWC proposes using the second-order approximation of this KL-divergence:
     \begin{equation}
    	K L\left(p\left(\boldsymbol{\theta} \mid \mathcal{D}_{1: i}\right) \| p\left(\boldsymbol{\theta} \mid \mathcal{D}_{1: i-1}\right)\right) \approx \frac{1}{2} \sum_j F_{j j}\left(\theta_j-\theta_{1: i-1, j}^*\right)^2	
     \end{equation}
	Here, $F$ refers to the empirical Fisher matrix, only the diagonal of which is used in the approximation.
	
    As the network is optimized using maximum likelihood estimation, the semantic transfer highly depends on the examples used during training and their similarity coefficient.  That is, dataset components with less similarity suffer more semantic loss. To avoid data-dependent optimization, LIME (Local Interpretable Model-Agnostic Explanations) \cite{ribeiro2016should, selvaraju2016grad} based continual learning where the important semantics are learned using weighted LIME scores in combination with EWC is proposed. 
	The contributions of this paper are:
	\begin{itemize} 
		\item A novel and general framework where EWC regularization is combined with model explainability to enhance the classification performance of any neural network is proposed.
		\item Further, LIME scores of miss-classified samples from the previous task are used as a weighting factor during model optimization to have better semantic transfer learning between tasks in an IL setting and to build more generalized models. 
	\end{itemize}
    Section \ref{proposed_framework} details the proposed approach of LIME-based weighted loss IL with Section \ref{experimental_setup} describing the experimental setup and elaborations of the experiments in Section \ref{results_discussion}.

\section{Background \& Proposed Framework}
    \label{proposed_framework}

    The existing literature (\cite{lipton2018mythos}) on Explainable AI (XAI) is extensive and focuses particularly on addressing the relationship between model output and its user. Much of this literature is exploratory in nature, as the definition of explainability, intertwined with interpretability, is loosely defined. However, recently some research has been conducted on the use of incorporating XAI and IL to create robust, reliable, non-human-in-the-loop AI systems \cite{ede2022explain, weber2022beyond}. The proposed solution utilizes recent XAI approaches introduced in \cite{ribeiro2016should, huawei2022xai} and incorporates them into an IL framework.
    
    \begin{figure}[!ht]
		\centering
		\includegraphics[width=0.9\linewidth]{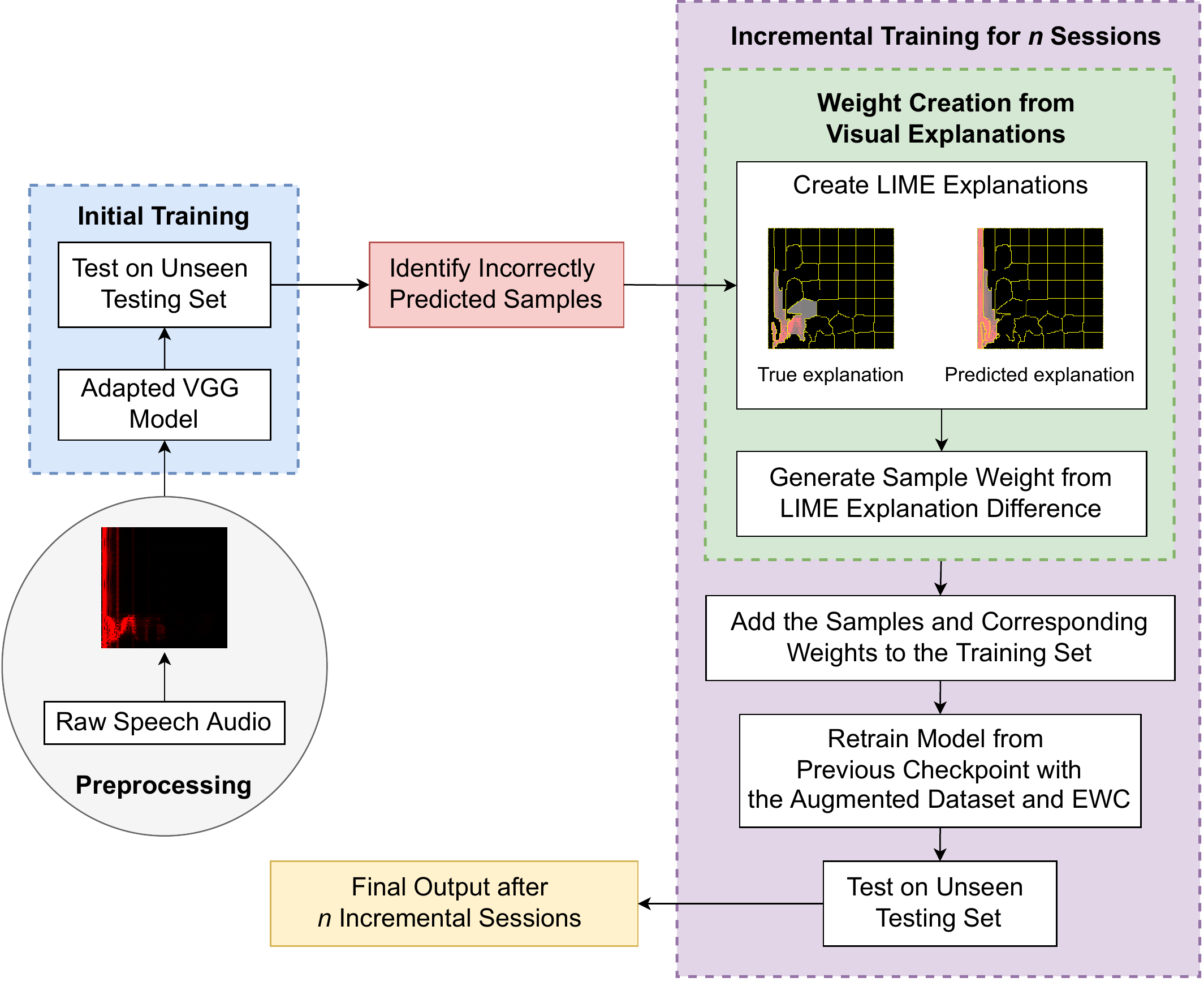}
		\caption{A high-level overview of the proposed training methodology that integrates explainable weight generation with EWC regularization to maintain model performance in sequential testing sets.}
		\label{fig: framework}
        \vspace{-2ex}
	\end{figure}
	
    The framework shown in Figure~\ref{fig: framework} aims to bridge this gap and provide a training methodology to enhance the classification accuracy for the KWS task in-midst of an explainable pipeline that can augment IL for the same task. The situation where adding new data to the training regime negatively impacts the learned distribution is known as catastrophic forgetting. In order to prevent this, \cite{kirkpatrick2017overcoming} proposed EWC, which forces the model to retain previous information on top of adding new data. However, although EWC acts as a regularizer to prevent catastrophic forgetting, it also limits the model to learn information from the new data. To address this, we propose using a weighted loss during model retraining where the weights from the samples come from the difference between LIME visuals of the true and predicted classes. As a result, the model will focus more on rectifying the incorrect predictions with higher weights, allowing the network to learn the new data better.

	\subsection{LIME Visualizations and Feature Scores}

    \begin{algorithm}
		\SetAlgoLined
		\SetKwInput{KwData}{Require}
		\KwData{$X_{in}, \theta^{t}$}
        $C =$ \textit{Cluster}$(X_{in})$ \\
        $var \gets [null], \sigma = 0.25$ \\
        \textbf{for} $i$ in $[1, 2, .... n]$ \textbf{do} \\
        $\quad V =$ \textit{Perturbation}$(C)$ \\
        $\quad var.append(V)$ \\
        \textbf{end for} \\
        $pred =$ \textit{Predict}$(\theta^{t}, var)$ \\
        $dist =$ \textit{Cosine Distance}$(X_{in}, var)$ \\
        $wt = \sqrt{e^{(-dist^2/\sigma^2)}}$ \\ 
        $reg =$ \textit{Linear Regression}$(var, pred, wt)$ \\
        $score = \operatorname{coeff}(reg)$ 
		\caption{Feature Score Generation Procedure with LIME}
		\label{algo:LIME}
	\end{algorithm}

\begin{figure}[!ht]
		\centering
		\includegraphics[width=1.0\linewidth]{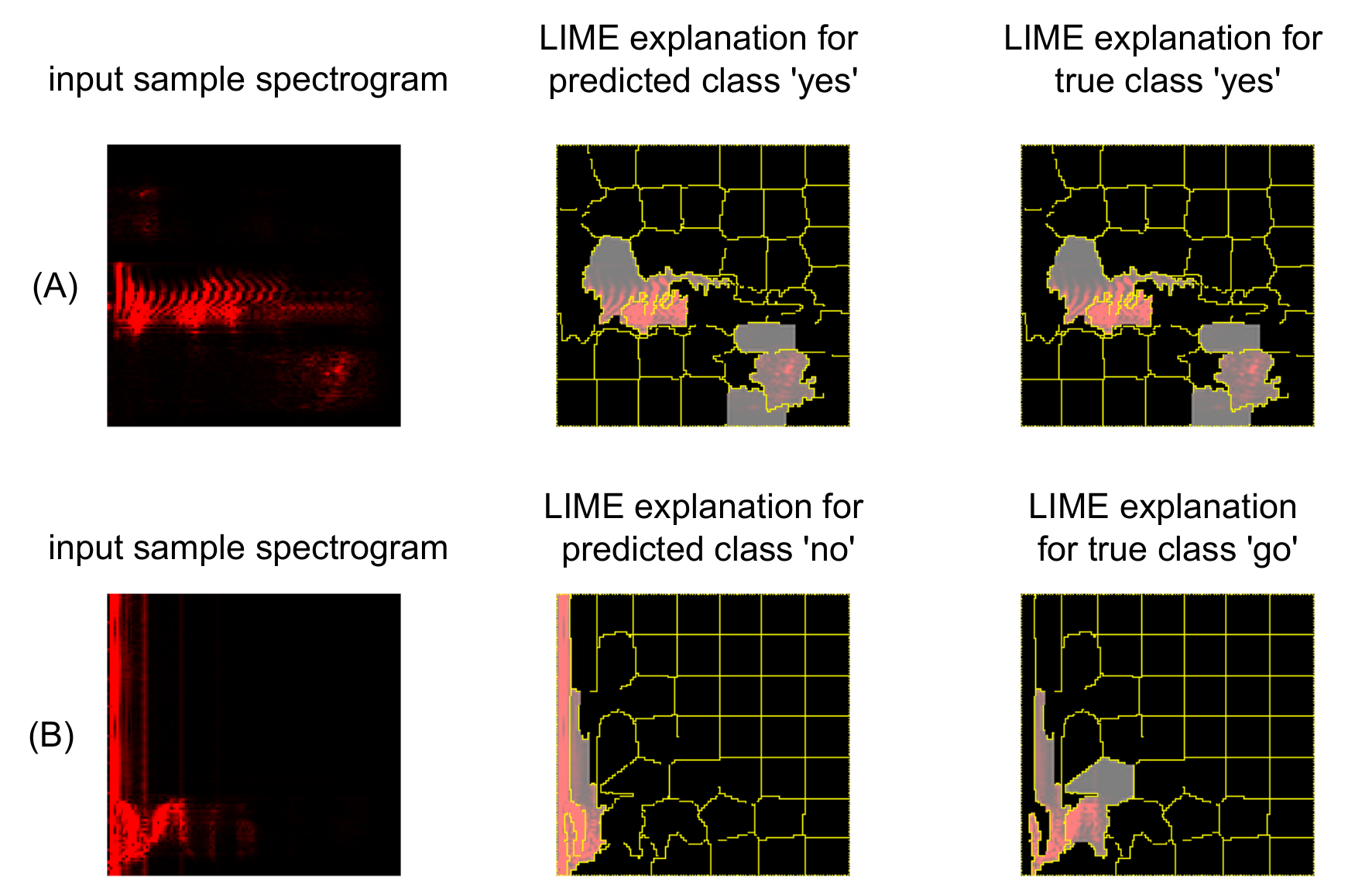}
		\caption{A qualitative representation of LIME explanations for an accurate classification (A) and a misclassification (B). (A) illustrates the five most influential \textit{slic} clusters (greyed out clusters) for making the prediction 'yes', which coincides with the LIME explanation for the true class. (B) demonstrates the two most influential \textit{slic} clusters for making the prediction 'no', which is different compared to the important \textit{slic} clusters of the true class 'go'.}
		\label{fig: lime_vis}
        \vspace{-3ex}
	\end{figure}

     Saliency map-based methods like GradCAM \cite{selvaraju2016grad}, ScoreCAM \cite{wang2020score}, etc., can provide explainable visuals of the heatmap overlayed on top of the original image according to either weight or gradient activation. However, in the context of enhancing the accuracy of the KWS task, determining the importance of segments within a spectrogram to isolate activity regions is highly important. To this end, the LIME-based visual explanations \cite{ribeiro2016should} to generate sample weight for the weighted loss is adopted in this work and detailed in Algorithm~\ref{algo:LIME}. Initially, the input is segmented with the \textit{slic} \cite{achanta2010slic} clustering algorithm. The clustered input then goes through a perturbation process where the different segments are turned on or off according to the binomial distribution. Each instance of the clustered input thus generated is referred to as a variation. Next, the trained model (for which the explanations are generated) is used to predict the classes of all these variations. Finally, the scores are created for the segments by fitting a linear regression classifier on the variations and their corresponding predictions, where the cosine distance between the variations and the original input act as weights. Since LIME explanations are based on segmentations created on the spectrogram, it tries to fit a linear classifier for generating an importance score for the segment itself. As a result, LIME provides a qualitative and a quantitative metric to explain the model prediction on an example as shown in Figure~\ref{fig: lime_vis} (A) and (B). 
     
 \subsection{Weighted Loss}
    In a traditional training setup, all samples are provided a similar influence/weight during loss generation. However, some samples are more difficult to predict than others. 
    More importantly, in an IL setting, it is imperative to learn the incoming data rigorously while preserving the already learned information from the distribution. In this context, using a weighted loss function will force the model to prioritize learning the new samples in back-propagation during the training. For example, if a sample is $X_i$, with the corresponding model to be $\theta(x_i,w)$ with $w$ parameter weights, and the sample loss to be $L(\hat{y_i},y)$ then the weighted loss function for $N$ batches in a feed-forward network is given by,

    

    \begin{equation}
        \begin{aligned}
            Loss = \frac{1}{N}\sum_{i=1}^{N} {w_iL(\hat{y_i},y)}
        \end{aligned}
        \label{eq:weighted_loss}
    \end{equation}
    
    The straightforward way to generate these above-mentioned sample weights would be to assign each incorrectly predicted sample with a high value. However, that process might not represent which samples need more focus than others. To address this, sample weights are generated from the explainable LIME visuals. The weights are considered to be the Euclidean distance between the LIME explanations for the true class $(E_t)$ and the predicted class $(E_p)$ as given by the following:
	
    \begin{equation}
        \begin{aligned}
            w = \sum_{i=1}^{i=n} {(E_p^i - E_t^i)^2}
        \end{aligned}
        \label{eq:lime_wt}
    \end{equation}	

    Here, $n$ denotes the number of segments. The choice behind using Euclidean distance is further justified through Table~\ref{tab:distance_metric}, where it is evident that Euclidean distance-based LIME weights outperform Manhattan and Cosine distance-based LIME weights in terms of classification accuracy.

    \begin{table}[htbp]
    \caption{Standard error of accuracy for different distance metrics over six separate runs}
    \label{tab:distance_metric}
    \begin{center}    
    \begin{tabular}{c|cl}
    \hline
    Distance Metric & \multicolumn{2}{c}{Accuracy (\%)}              \\ \hline
    Euclidean       & \multicolumn{2}{c}{\textbf{90.92 \textpm 0.23}} \\ \hline
    Manhattan       & \multicolumn{2}{c}{90.69 \textpm 0.28}          \\ \hline
    Cosine          & \multicolumn{2}{c}{90.65 \textpm 0.24}          \\ \hline
    \end{tabular}
    \end{center} 
    \vspace{-2ex}
    \end{table}
    \vspace{-2ex}
 \subsection{Elastic Weight Consolidation (EWC)}

    To address catastrophic forgetting in an IL setting, it is imperative to have some regularization (usually L1 or L2) during training to maintain information from the previous  distributions. However, as demonstrated in \cite{kirkpatrick2017overcoming}, L1 or L2 normalization most often constrains each weight with the same coefficient, meaning the model can only remember the previous task at the expense of not learning the new one. EWC provides an alternative to this predicament where learning is slowed down on certain weights based on their importance to the previous distribution with a quadratic penalty on the loss, as shown in Equation~\ref{eq:EWC}. This scenario further enhances the learning of both tasks and maintains the model performance on each separately. For example, if the loss for the current task is $L_{curr}$, then EWC regularization is given by the following:

    \begin{equation}
        \begin{aligned}
            Loss = L_{curr} + \sum_{i} {\frac{\lambda}{2}F_i(\theta_i - \theta_{A,i}^*)^2 }
        \end{aligned}
        \label{eq:EWC}
    \end{equation}	
    
    Here, $\theta_{A,i}$ are the parameters from the previous task, and $\theta_i$ are the parameters from the current task. $F_i$ represents the parameters of the previous task's Fisher Information Matrix (FIM). $\lambda$ controls the amount of EWC regularization applied to the current loss. A lower $\lambda$ value favors learning the new task more at the expense of forgetting the previous task and vice versa.
	 
	\begin{algorithm}
		\SetAlgoLined
		\SetKwInput{KwData}{Require}
		\KwData{$D^{m}, D^{v}, D^{t}$}
		\textit{Initial Training :} \\
		1: Initialize model $\theta^{0}$ \\
		2: $\theta^{m}=\operatorname{train}\left(D^{m}, \theta^{0}, D^{v}\right)$ \\
		3: $A c c^{m}=\operatorname{eval}\left(\theta^{m}, D^{t}\right)$ \\
		\textit{Generate LIME weights} \\
		4: \textbf{for} $D^{v_i}$ in $[D^{v_1}, D^{v_2}, .... D^{v_n}]$ \textbf{do} \\
        5: $\quad D^{I_i} = \{(d,y) \in D^{v_i}|pred(\theta^m,d) \neq y \}$\\
        6: $\quad LIME\_pred^{I_i} = lime_p(\theta^m, D^{I_i})$ \\
        7: $\quad LIME\_true^{I_i} = lime_t(\theta^m, D^{I_i})$ \\
        8: $\quad w^{I_i} = l\_wt(LIME\_pred^{I_i}, LIME\_true^{I_i})$ \\
        9: \textbf{end for} \\
        \textit{Incremental Training :} \\
        10: Initialize $D = D^m, \theta^0 = \theta^m, w = 1$ \\
        11: \textbf{for} $D^{I_i}$ in $[D^{I_1}, D^{v_2}, .... D^{I_n}]$ \textbf{do} \\
        12: $\quad$Add new data: $D_{new} = D \cup D^{I_i}, w = w \cup w^{I_i}$ \\
        13: $\quad FIM = gen\_fim(\theta^{i-1}$, 5\% of $D_{new})$ \\
        14: $\quad \theta^i = EWC\_trn(\theta^{i-1}, D_{new}, w, D^{v_i}, FIM)$ \\
        15: $\quad Acc^i = eval(\theta^i, D^t)$ \\
        16: \textbf{end for}
		\caption{Incremental Training Procedure with LIME-based Weighted Loss and EWC.}
		\label{algo:inc}
	\end{algorithm}
	%
	%
\section{Experimental Setup}
    \label{experimental_setup}

    The Google Speech Commands dataset \cite{warden2018speech} forms the basis of our experimental protocol. The process starts with normalizing all audio files to 16,000 samples through zero-padding. Following this, we generate spectrograms from these files, thereby considerably diminishing the computational load for subsequent deployment of Deep Neural Networks (DNNs). The derived spectrograms further facilitate LIME-based explanations that serve as valuable feedback in the learning mechanism. For this study, a VGG-like architecture is leveraged, albeit with only four blocks and lesser filters across all layers compared to the conventional VGG structure. This modified VGG structure is henceforth referenced as the study's adapted VGG architecture (illustrated in Equation~\ref{eq:adapted_VGG}). In this notation, 2$\times(8\,Conv_{3 \times 3})$ denotes 2 consecutive convolutional layers with 8 filters and a 3$\times$3 kernel shape, $32\, pool_{2 \times 2}$ signifies a 2$\times$2 maxpooling layer with 32 filters, and $1000\, FC$ represents a fully-connected layer hosting 1000 neurons.

    \vspace{-2.5ex}
    \begin{equation} 
    \footnotesize
    \begin{aligned}
    (in)-2\times(8\,Conv_{3 \times 3})\!- \!2\times(16\, Conv_{3 \times 3})\!-3\times (32\, Conv_{3 \times 3})\!- \\
    (32\, pool_{2 \times 2})\!-3\times (64\, Conv_{3 \times 3})\!-(64\, pool_{2 \times 2})\!-(1000\, FC)\!-(out)
    \end{aligned}
    \label{eq:adapted_VGG}
    \end{equation}
    
    The comprehensive training approach is depicted in Figure~\ref{fig: framework} and elaborated in Algorithm~\ref{algo:inc}. To ensure viable sequential retraining, we partition the data into training, validation, and test sets at 80\%, 10\%, and 10\% respectively, strictly adhering to speaker information for the division. This ensures complete containment of a speaker's utterances within a single set, thus preventing any potential leakage of samples between sets. For the purpose of retraining, misclassified samples from the validation sets are moved into the training set. These samples are weighted based on the Euclidean distance between the LIME explanation of the predicted class ($lime_p$) and the actual class ($lime_t$), as denoted in line 8 of Algorithm~\ref{algo:inc}. The stepwise training procedure then unfolds by introducing these misclassified samples and corresponding LIME weights into the training data $D_m$. This incremental methodology continues for all sequential training iterations. The choice of this strategy arises from the notion that the user may not always have access to previously unseen test data or validation data. It also offers the advantage of significant reduction in training time, given that only the misclassified samples are considered for retraining, instead of the entire validation set. To formulate the Fisher Information Matrix (FIM) during EWC-based training, we randomly select 5\% of the correctly predicted samples from the expanded training set in each session. In all sessions, the retraining commences utilizing the weights inherited from the previous session, leveraging the Adam optimizer (with a learning rate of 0.001) and a batch size of 512.

	\section{Results \& Discussion}
    \label{results_discussion}

     \begin{figure*}[!ht]
		\centering
		\includegraphics[width=1.0\linewidth]{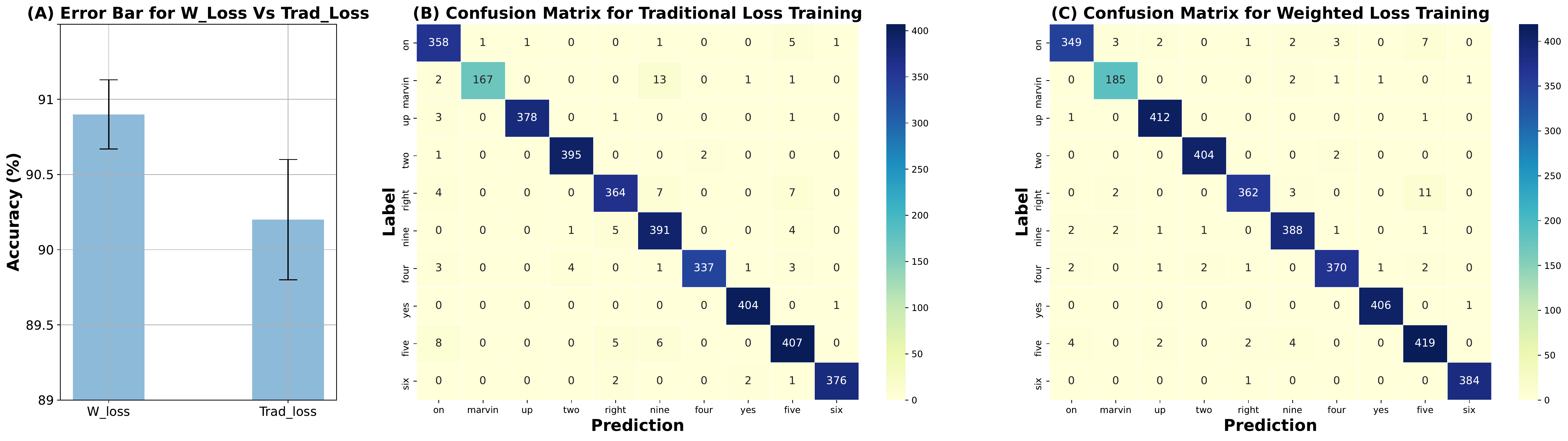}
		\caption{(A), (B), and (C) correspond to the error bar for two training methods over six runs, confusion matrix for weighted loss-based training, and traditional loss-based training, respectively, on the complete dataset. Only the first ten classes are represented in the confusion matrix figures. The rest of the classes follow a similar trend as well.}
		\label{fig: cm}
	\end{figure*}
 
\begin{figure}[!ht]
		\centering
		\includegraphics[width=1.0\linewidth]{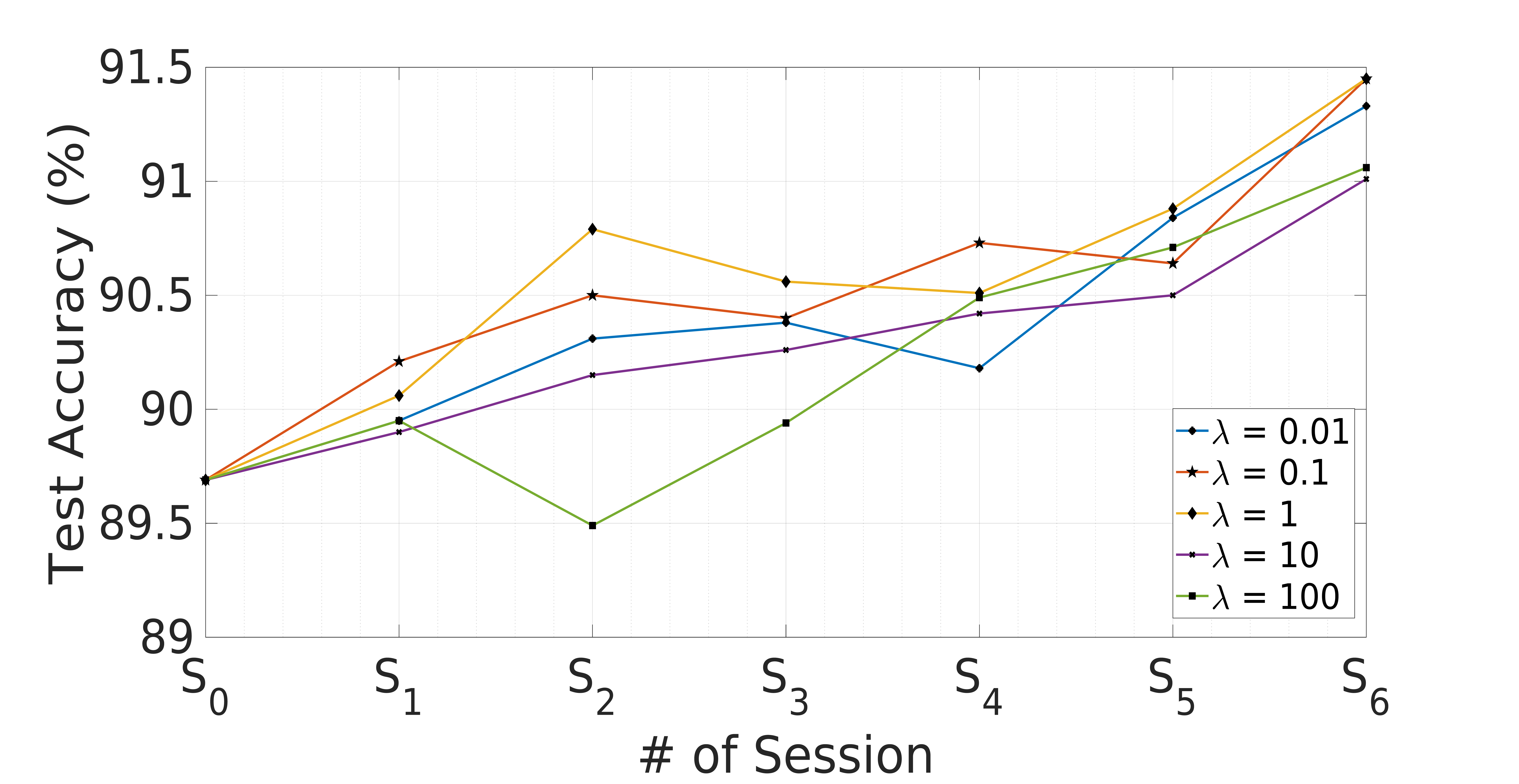}
		\caption{The experimental insight allows us to conclude that $\lambda = 1$ provides the best performance in terms of accuracy throughout the six different sessions of IL.}
		\label{fig: lambda_exp}
	\end{figure}

    We perform six separate runs on the complete dataset for weighted loss training and regular loss training to create an analogy between the two methods. The weighted loss method results in 90.9\% Top-1 accuracy, which is around 1\% better than regular loss-based training (90.2\%). Figure~\ref{fig: cm} shows the performance of the first ten classes during testing in the form of confusion matrices. The weighted loss training method adopted in this work performs better and gets more predictions right in most classes than traditional loss-based training, proving that our custom loss enhances performance. To further accommodate EWC according to Equation~\ref{eq:EWC}, we aim to choose the optimal parameter ($\lambda$), which controls the influence of EWC regularization while generating the loss. A higher $\lambda$ value allows aggressive learning of the newer task with the caveat of forgetting the older ones. Hence we experiment with six identical EWC and weighted loss-based IL setups for six different $\lambda$ values. As per the line graph shown in Figure~\ref{fig: lambda_exp}, it is apparent that $\lambda = 1$ provides us with the most stable performance in terms of test set accuracy throughout all six sessions of IL.

    \begin{table}[htbp]
    \caption{Standard Error Comparison among traditional loss, weighted loss and weighted loss with EWC for 16 stage incremental learning over three separate runs}
    \label{tab:compare}
    \begin{center}
    \begin{tabular}{c|c|c|c}
    \hline
    \# of Sess. & Trad. Loss         & W\_Loss            & W\_Loss + EWC      \\ \hline
    1           & 89.37 \textpm 0.75 & 90.17 \textpm 0.71 & 90.43 \textpm 1.08 \\ \hline
    4           & 89.90 \textpm 0.56 & 90.20 \textpm 0.78 & 91.00 \textpm 0.95 \\ \hline
    8           & 89.73 \textpm 0.51 & 90.13 \textpm 0.50 & \textbf{91.07 \textpm 0.49} \\ \hline
    12          & 89.93 \textpm 0.75 & 90.03 \textpm 0.51 & 90.63 \textpm 0.70 \\ \hline
    16          & 89.57 \textpm 0.60 & 90.03 \textpm 0.46 & \textbf{90.87 \textpm 0.65} \\ \hline
    \end{tabular}
    \end{center}
    \end{table}
    
    Table~\ref{tab:compare} compares the performance of the EWC-based weighted loss IL to traditional IL and stand-alone weighted loss IL, respectively. The 16-stage IL experiment suggests that the weighted loss incremental training performs better than traditional incremental training throughout all the sessions. And, when coupled with EWC regularization, the accuracy is better for EWC-based weighted loss IL than the stand-alone weighted loss IL, further providing evidence that our proposed method retains previous information better than traditional IL setups. Moreover, we see the test accuracy follows a reducing trend for all three methods as more incremental stages are trained. This effect is the result of some information from the earlier stages being lost. Finally, our proposed method has an improvement of at least 0.5-1.5\% over traditional loss based IL setup for all stages of IL and even with higher training stages our method maintains better performance compared to both traditional and weighted loss based IL setups.

	\section{Conclusion}
    \label{conclusion}

     This work aims to take advantage of the LIME visuals of a model prediction and use these explainable insights to improve the performance of DNNs. By differentiating the explanations for actual and predicted classes, proportional weights are generated for false predictions and used during the consequent IL stage retraining. On top of it, the enhancement brought upon by EWC regularization to maintain neural network performance across IL scenarios is explored. The experiments suggest that the weighted loss training coupled with EWC provides at least 0.5-1.5\% improvement on all IL stages over the traditional loss-based IL. Our future work will expand the scope of this methodology to include weights from explainable visuals in the context of metric learning to further bolster the classification network's performance.

	\bibliographystyle{./IEEEtran}
	\bibliography{./ref.bib}

\end{document}